\documentclass[11pt,a4paper]{article}
\usepackage{authblk}
\usepackage[hyphens]{url}  

\usepackage{eacl2021}  
\usepackage{times}  
\usepackage{helvet} 
\usepackage{courier}  
\usepackage{graphicx} 

\usepackage{latexsym}

\usepackage{booktabs}
\usepackage{microtype}
\usepackage{float}
\usepackage{amsfonts}
\usepackage{makecell}

\usepackage{natbib}
\usepackage{bibentry}
\usepackage{tabularx}
\usepackage{authblk}
\usepackage{amsmath}
\usepackage{cleveref}

\aclfinalcopy 

\title{Visual Grounding Strategies for Text-Only Natural Language Processing}

\author[ ]{Damien Sileo}

\affil[ ]{Department of Computer Science (CS)}
\affil[ ]{KU Leuven}
\affil[ ]{damien.sileo@kuleuven.be}

\begin{document}

\maketitle

\begin{abstract}
Visual grounding is a promising path toward more robust and accurate Natural Language Processing (NLP) models. Many multimodal extensions of BERT (e.g., VideoBERT, LXMERT, VL-BERT) allow a joint modeling of texts and images that lead to state-of-the-art results on multimodal tasks such as Visual Question Answering. Here, we leverage multimodal modeling for purely textual tasks (language modeling and classification) with the expectation that the multimodal pretraining provides a grounding that can improve text processing accuracy. We propose possible strategies in this respect. A first type of strategy, referred to as {\it transferred grounding} consists in applying multimodal models to text-only tasks using a placeholder to replace image input. The second one, which we call {\it associative grounding}, harnesses image retrieval to match texts with related images during both pretraining and text-only downstream tasks. We draw further distinctions into both strategies and then compare them according to their impact on language modeling and commonsense-related downstream tasks, showing improvement over text-only baselines. 

\end{abstract}
\section{Introduction}

Representation of text with transferable encoding is a central task of artificial intelligence. Transfer from larger and larger transformer-based text encoders trained on masked language modeling has become a standard way to achieve state-of-the-art results. Progress is shown in tasks such as natural language inference and semantic similarity estimation when evaluated on natural language understanding benchmark datasets \citep{kaplan2020scaling} such as GLUE and SuperGLUE \citep{wang2019superglue}.
However, these scores do not tell the whole story. Firstly, marginally better benchmark scores can come at the price of impractical GPU requirements. Secondly, super-human scores can be obtained by exploiting spurious dataset-specific correlations instead of more generalizable reasoning \citep{niven-kao-2019-probing}. Mastering commonsense reasoning is regarded as a requirement for "true" language understanding \citep{bisk2020experience}, and grounding representations of natural language on other modalities such a visual perception is a privileged strategy in that endeavor. Since the meaning of language stems from the physical world, visual grounding\footnote{Here, we define visual grounding as learning language representations from explicit visual associations.} is a valuable way to guide the training of NLP models. Thus, we hypothesize that text encoders, even already pretrained on a massive amount of text-only data, can be improved by a further multimodal pretraining stage. 

Many visuolinguistic transformer architectures such as LXMERT \citep{tan-bansal-2019-lxmert} VilBERT \citep{lu2019vilbert} or VL-BERT \citep{su2019vl} have been proposed to augment BERT \citep{devlin-etal-2019-bert} with joint text-image understanding in order to tackle multimodal tasks including visual question answering and image retrieval, each leading to substantial performance improvement over the previous state of the art. To that end, these models generalize BERT's Masked Language Modeling (MLM) objective to a multimodal setting. More specifically, they perform the MLM task on image captions while allowing the model to use features of the paired image\footnote{An image is represented as a sequence of regions features extracted with a Convolutional Neural Network.}, and also perform image modeling while letting the model attend to textual features. Nevertheless, it is not clear whether joint pretraining is better than text-only pretraining when transferring to text-only tasks (e.g., language modeling, text classification, similarity estimation) and how it should be done. In this work, we address this question by breaking down the strategies toward that endeavor.
 
 A first proposed way to leverage visuolinguistic models for textual tasks is to perform the aforementioned joint modeling pretraining and then fine-tuning on text-only tasks while using a placeholder as image input. However, this only trains the text encoder on the caption domain (which is different from the wider domain usually used to train text encoders) and the model is not exposed to any training example without images. To alleviate this problem, it is possible to combine joint masked modeling on captioned images with unimodal MLM on a broader domain text-only corpus during pretraining by associating an image placeholder with the text-only examples. We call this strategy {\it transferred grounding}.
 
 We also propose another technique named {\it associative} grounding where for each textual input, an association module retrieves the most relevant images from a large external images collection. The text encoder parameters can then harness these externally provided images for better text understanding instead of having to internally model them.


Since we are bringing an additional computational cost by using images and not only pure text, we will strive to reduce the incurred additional memory usage in our models.

The goal of this paper is to identify and compare techniques that bridge visuolinguistic multimodal modeling with text-only tasks. To that aim, our contributions are the following: (i) $4$ visuolinguistic grounding strategies for text-only tasks transfer; (ii) An evaluation of the size/accuracy trade-off of image representation to reduce the memory usage of multimodal transformers; and (iii) A systematic comparison of the proposed strategies for transformers on English text-only tasks.

\section{Grounding strategies }

\begin{figure}[]
  \centering
  \vspace{-0cm}
\includegraphics[width =0.47\textwidth, page=3,]{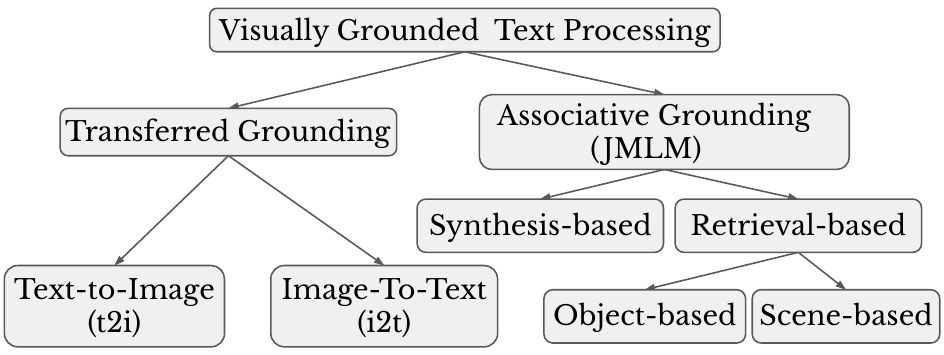}
  \caption{A categorization of the proposed grounding strategies.}
\label{fig:categorization} 
\end{figure}

In this section we present the two main grounding strategies, that is \textit{transferred grounding} and \textit{associative grounding}, of which Figure \ref{fig:categorization} shows an overview. We first describe the multimodal model architecture used in both grounding strategies.

\subsection{Multimodal model structure}

\begin{figure*}[]
 \centering
 \frame{
\includegraphics[width =0.9\textwidth, page=2]{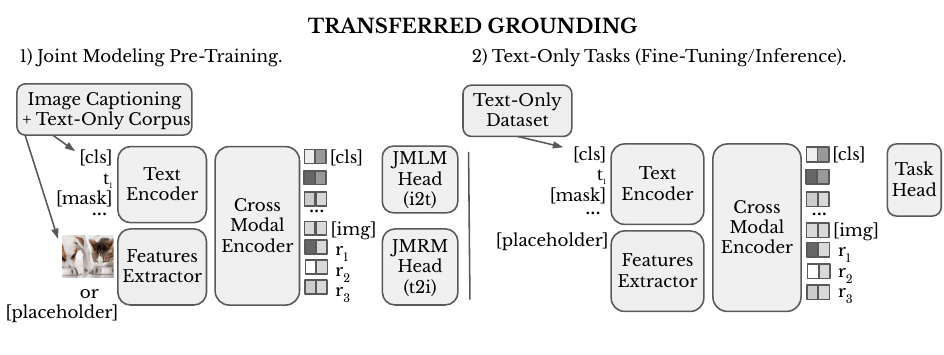}
}
  \caption{Overview of the Transferred grounding strategy.}
\label{fig:transfergrounding} 
\end{figure*}

We rely on a setup shared by several multimodal extensions of BERT \citep{tan-bansal-2019-lxmert, su2019vl, lu2019vilbert}.
The text encoding 
takes place in two stages.
The output of the text unimodal encoder (e.g., BERT) provides an intermediate representation of the text input. The visual input has a vector representation (typically CNN-extracted embedding possibly contextualized with a transformer model). Then, both are fed to a cross-modal encoder that learns to compose the modalities in order to perform visuolinguistic modeling tasks. This multimodal model structure is illustrated in both Figures \ref{fig:transfergrounding} and \ref{fig:associativegrounding}.

 This architecture could directly be used in order to perform downstream tasks. However, a multimodal pretraining, for instance through masked modeling, is key to achieve high performance \citep{tan-bansal-2019-lxmert}.

\subsection{Masked modeling of multimodal features}
In our grounding strategies we rely on masking. In Joint Masked Language Modeling (JMLM), random tokens of the input text $t$ are replaced by a $[\texttt{masked}]$ token. A softmax JMLM head on top of the cross-modal encoder has to predict the original value of the masked tokens through cross-entropy ($\text{H}$) loss minimization. The cross modal encoder is thus incentivized to use the visual input regions $r$ when they are relevant for the masked tokens prediction. Joint Masked Region Modeling (JMRM) is the visual counterpart of JMLM. Here, image region features are masked and have to be predicted, for example, through a $\text{L}_p$ loss minimization, based on both the text and non-masked visual cues. The corresponding losses are the following: 
\[
\mathcal{L}_{\text{JMLM}}=\sum_{k=1}^{|t|}m_k \text{H}(t_k, \hat{t}_{k}(\bar{t},r))
\]
\[
\mathcal{L}_{\text{JMRM}}=\sum_{k=1}^{|r|}m_k \text{L}_p(r_k, \hat{r_k}(t,\bar{r}))
\]
where $m_k=1$ when a token/region is masked, $\bar{t}/ \bar{r}$ represents the masked tokens/regions sequence, $\hat{t}$ the predicted probability distributions, and $\hat{r}$ the predicted region features.

\subsection{Transferred grounding}

In \textit{transferred grounding} the masked modeling of multimodal features is used in pretraining as shown in Figure \ref{fig:transfergrounding}. 

The corpus used for pretraining contains images paired with captions, which can also be extended with another text-only corpus. Adding examples from a text-only corpus allows the encoder to be exposed to a wider domain of text during pretraining. Since no image is available for these examples, we replace the image features by a single trainable embedding $h_{\text{[placeholder]}} \in \mathbb{R}^d$ where $d$ is the dimension of the cross-modal encoder inputs.

Cross-modal prediction can occur in the two following directions, each providing a different way to ground language understanding.

\subsubsection{Transfer from Text-To-Image prediction (t2i)}
Pretraining a model to perform JMRM incentivizes the text representations to abstract the visual knowledge involved in visual region modeling. This provides a form of grounding that could be useful for textual downstream tasks. Here the model learns to predict masked image aspects from text, which might help language understanding by visually imagining the language content.

\subsubsection{Transfer from Image-To-Text prediction (i2t)}
A different way to perform pretraining is to perform JMLM in the presence of visual input. The model can learn to use visual information for textual modeling, thus developing useful abstractions. We hypothesize that these abstractions can help text-only tasks when the visual input they relied on is missing.

\subsubsection{Text-only tasks}

Once pretrained, the multimodal architecture can be applied to text-only tasks. To do so, the placeholder features $h_{\text{[placeholder]}}$ are used to replace the missing visual input.

\begin{figure*}[]
 \centering
 \frame{
\includegraphics[width =0.9\textwidth]{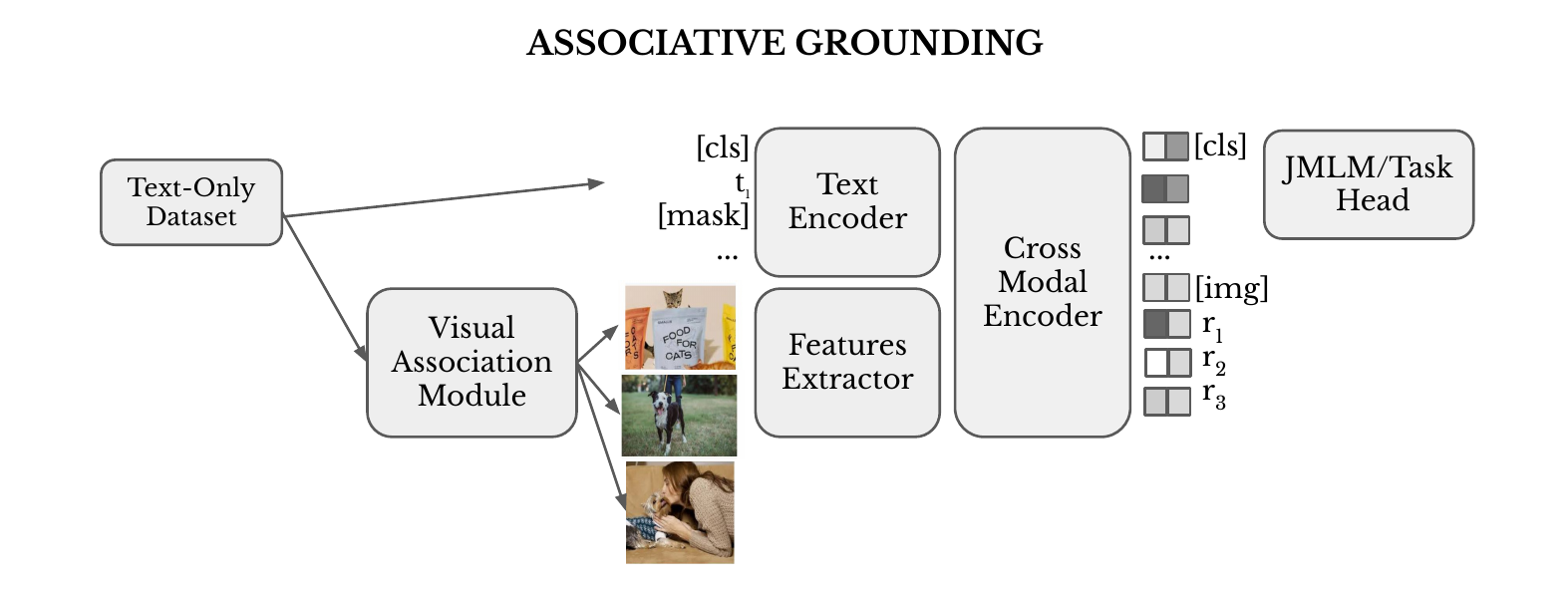}
}
  \caption{Overview of the Associative grounding strategy.}
\label{fig:associativegrounding} 
\end{figure*}

\subsection{Associative Grounding}

In an alternative grounding strategy, called \textit {associative grounding}, we do not rely on a text corpus that is a priori paired with images, but the pairing is part of the model. 
As seen in Figure \ref{fig:associativegrounding}, for each text input\footnote{During JMLM pretraining, the masked text is used to find associations. This to prevent trivial prediction of the masked text tokens from their presence in the images.}
 $t$ (e.g., sentence or paragraph), a visual association module retrieves the most relevant images from an archive of images.

Here, we instantiate the visual association module in the following way.
A query encoder provides representation $q_t$ for the masked input text $t$ using the cross-modal encoder described above, and a relevance metric R($q_t$,$k_i$) such as the cosine similarity identifies the $K$ most relevant images in a collection $M$ of images $i$, which are encoded into key vector representations $\{k_i, i \in M \}$. 

Associated images can be visual scenes that match the input text as a whole. However, the space of situations matching the content of the input text is immense, and in many cases, the most relevant images in $M$ will only be loosely associated to the textual input.
Decomposing the situation evoked in the text into objects that play an important role can help narrowing down the space of possible relevant images, thus leading to closer associations, albeit more partial. We call the first substrategy {\it scene-based} and the second one {\it object-based}.

\subsubsection{Scene-based association}
Here we propose to use textual captions to index images.
Each input text $t$ is then associated with the $K$ most similar images, according to the relevance computation of a caption given the input text. Of course, retrieved images may not be sufficiently relevant, and it is not clear whether some sentences can be illustrated by existing images. But we hypothesize that the cross-modal encoder of the captions is able to ignore non-relevant information. \citet{kiela-etal-2014-improving} have shown that image retrieval systems map abstract words to high dispersion results, that is, retrieved images are different to each other. This dispersion could be informative for the cross-modal encoder. Besides, concrete images could also ground more abstract concepts \citep{LakoffJohnson80}; a very simple illustration of this is that images captioned with references to {\it sadness} tend to be darker than those containing the word {\it happiness}.

\subsubsection{Object-based associations}

Another possibility is to rely on objects mentioned in the text. Here, we extract all common nouns in the input text and perform a Gaussian Mixture Clustering to extract $\kappa \leq K$ clusters among the noun representations to find the key concepts. For each centroid, the closest noun is chosen as a representative. A noun-indexed database, here ImageNet22K \citep{deng2009imagenet}, can be used to map these nouns to images.

The two association systems used in the experiments are illustrated in Figure \ref{fig:matching} in the appendix.

\subsubsection{Image synthesis}
It would also be possible to synthesize images that represent the input text with a dedicated model. We experimented with the {\it DeepAI} system\footnote{\url{https://deepai.org/machine-learning-model/text2img}} and as shown in Figure \ref{fig:associativegrounding}, the results suggests that more work is needed to get conclusive results. Consequently, we do not further consider image synthesis here.

\section{Experiments}

We evaluate the proposed grounding strategies in language modeling and downstream tasks.

\subsection{Text-only corpus}

Following \citep{devlin-etal-2019-bert}, we use a combination of Wikipedia English pages and the BookCorpus \citep{Zhu_2015_ICCV} as language modeling corpus. We extract the body of pages from the Wikipedia pages with at least 20 views in the 2013 dump\footnote{Available at \url{https://storage.googleapis.com/lateral-datadumps/wikipedia_utf8_filtered_20pageviews.csv}}, totaling 463k documents, and we sample the same number of passages from the BookCorpus. We call this combination Wiki-BC.
We select $90\%$ of the pages/passages as training data, $5\%$ for validation and $5\%$ for test.

\subsection{Visuolinguistic corpus}

For the {\bf scene-based associations}, we populate the scenes bank $M$, with the combination of two image captioning datasets. The Stony Brook University corpus (SBU) \citep{Ordonez:2011:im2text} is composed of 860k images from the Flickr website, filtered to ensure that the caption literally describes the image. ConceptualCaptions (CC) \citep{sharma-etal-2018-conceptual} gathers 3M images that come from web pages that also were heuristically filtered. 
We use a $90/5/5 \% $ train/validation/test split for SBU and the standard split for CC.

When performing {\bf object-based associations}, $M$ is populated with ImageNet \citep{deng2009imagenet}. We keep the $15k$ synsets that are associated to at least $10$ images, and we randomly sample the images when more are available. Each image is indexed with the average of the lemmas embeddings of its synset, averaged with the embedding of the synset definition.

When performing {\bf transferred grounding}, we concatenate the captioned images (CC-SBU) with our text-only corpus. We sample CC-SBU so that its size matches the text only corpus size (840k passages) to balance the train set. Since our end-goal is text-only processing, we do not use the captioned images in the validation and test set but only in the train set.
 
In order to test image representations and language modeling on literal text, we also run experiments on the COCO \citep{lin2014microsoft} dataset made of 400k image/captions pairs. When we do so, we discard the original images when using the associative setup, but we always keep the images in the transferred setup.


\subsection{Setup and hyperparameters}

 We build upon the LXMERT \citep{tan-bansal-2019-lxmert} code and experimental setup, and rely on its hyperparameters values because of the state-of-the-art results and code availability of this model.
Our text encoder is an Albert model pretrained from {\it albert-base-v2} checkpoint \citep{lan2019albert}. Its weights are kept fixed during pretraining and fine-tuned during downstream task training. We use a different copy of this Albert model as the cross-modal encoder initialization and always keep its weights trainable.

In this work, our key/query encoder is a continuous bag of words.
 We use fastText \citep{mikolov2018advances} embeddings\footnote{We use the CommonCrawl  version \url{https://fasttext.cc/docs/en/english-vectors.html.} and discard stop-words with NLTK.} which are competitive with state-of-the-art models on semantic similarity tasks \citep{reimers-2019-sentence-bert}. Queries and keys are matched according to cosine distance with the Faiss \citep{JDH17} library.

Following LXMERT, each image is represented with a sequence $N$ region features. A linear projection maps the region features to the input space of the cross-modal encoder. We also provide $K$ rank embeddings which are added to the cross-modal encoder input image representation. Input text is lower-cased and token sequence length is clipped to $64$.  We use the Adam \citep{kingma2014adam} optimizer with a batch size of $32$ and a learning rate of $10^{-4}$ for downstream tasks until convergence on the validation accuracy, with a maximum of $4$ epochs. When we retrieve $K=16$ images \footnote{We have tested values $\{1,4,8,16,24\}$ and the lowest Wiki-BC validation perplexity for both object and scene based setups is 16.} with object-based associations, we use $\kappa=8$ clusters\footnote{We capped this value to 8 since some short texts do not refer to many objects.}.
Following  \citep{tan-bansal-2019-lxmert}, we mask $15\%$ of text tokens during the JMLM pretraining; and we use a linear regression head with $\text{L}_1$ regularization for JMRM.

We first compare the expressivity of the image representations in function of their size, then delve into the comparison of grounding strategies, according to JMLM pretraining and downstream tasks scores.

\subsection{Efficient image representation}

We want to be able to provide $K > 1$ images to the cross-modal encoder. However, the cross-modal encoder memory usage scales quadratically with the number of regions $K.N$, so we will evaluate to what extent $N\gg1$ is necessary. Previous work use object detection on a single image to find $36$ Regions of Interest (RoI) and represents them by ResNet-50 \citep{he2016deep} features with 2048 dimensions. We run the JMLM pretraining setup with several image representations. We evaluate two representations: (1) the previously used ResNet region representations provided by  \citep{tan-bansal-2019-lxmert} with various values of $N$; (2)  Single representation of the whole image, using EfficientNet-B1 Noisy Student \citep{Xie_2020_CVPR} pre-logit max-pooled features. We finetune neither of them.

We use the COCO dataset for this evaluation because its annotations are manual and not heuristically filtered, and because the image features used by \citep{tan-bansal-2019-lxmert} are publicly available. We evaluate the image  representations with JMLM pretraining perplexity.
Table \ref{tab:coco} shows the results of this experiment. Overall, the presence of image features ($N>0$) significantly improves the JMLM performance. While the standard FastRCNN\citep{girshickICCV15fastrcnn}+ResNet with 16 objects representation achieves the best perplexity (even better than with 36 features, which can be attributed to a reduced over-fitting), while this performance breaks down with few recognized objects. The performance of EfficientNet representations using the whole image as a single region is competitive while being much lighter. Consequently, in the experiments described in the following section we represent images with EfficientNet-B1 features and $N=1$ region per image.

\begin{table}
\centering
\small
\begin{tabular}{rll}
\toprule
  N &           Features & COCO perplexity \\
\midrule
  0 &                  - &            8.4 \\
  1 &  EfficientNet-1280 &            4.6 \\
  8 &        ResNet-2048 &            4.6 \\
 16 &        ResNet-2048 &   \textbf{4.0} \\
 36 &        ResNet-2048 &            4.2 \\
\bottomrule
\end{tabular}

\caption{The effect of image representations on cross-modal JMLM perplexity on the COCO dataset with $K=1$ image per example; $N$ is the number of visual regions per image.}
\label{tab:coco}
\end{table}

\section{Empirical comparison of grounding strategies }

\subsection{JMLM training}

We run the pretraining for each strategy on the Wiki-BC corpus, and also run a separate pretraining on the COCO corpus in order to assess language modeling performance on a literal, descriptive domain. Table \ref{tab:pretraining}  reports the influence of the grounding strategies on test perplexities. \citep{Zhang2020Neural} is an image retrieval system ranking the images according to the number of text keywords they contain. It was originally used to improve Neural Machine translation (see Related Work).
On the COCO dataset, using the original image paired with the text captions leads to the lowest perplexity, but surprisingly, using $K=16$ associated object or scenes leads to a competitive perplexity. This suggests than the many distantly related images can be a good substitution to a canonically associated image, even though associating $K=1$ object seems to be a too noisy signal.

\begin{table}
\begin{footnotesize}
\setlength\tabcolsep{4pt}
\centering

\begin{tabular}{llll}
\toprule
                            Strategy & K &  \thead[l]{Wiki-BC\\ ppl.} &    \thead[l]{ COCO\\ ppl.} \\
\midrule
                        No Grounding &        0 &           8.7 &           7.6 \\
                     Transferred i2t &      0/1 &          30.0 &  \textbf{4.6} \\
                     Transferred t2i &      0/1 &          41.6 &         - \\
                 Transferred t2i+i2t &      0/1 &          25.3 &         - \\
 Associative \citep{Zhang2020Neural} &       16 &           8.9 &           8.6 \\
           Associative (scene-based) &       16 &  \textbf{8.5} &  \textbf{4.6} \\
           Associative (scene-based) &        1 &           9.1 &           7.3 \\
          Associative (object-based) &        1 &           9.0 &          35.6 \\
          Associative (object-based) &       16 &           8.6 &           4.7 \\
\bottomrule
\end{tabular}

\caption{JMLM test perplexity on the Wiki-BC+CC-SBU dataset and on the COCO captions. On the COCO dataset, COCO images are used at test time for the transferred i2t strategy. On the Wiki-BC dataset, the image placeholder is used at test time for the transferred strategies while associative strategies always use associated images.
}
\label{tab:pretraining}
\end{footnotesize}
\end{table}
The results on the Wiki-BC dataset reveal a similar pattern. Here, the transferred strategies cannot use images at test time and suffer from a domain shift which leads to inferior results. However, the associative strategies and $K=16$ lead to the lowest perplexity even though the gap between ungrounded and grounded methods is small than with the COCO dataset due to abstractness of the open-domain texts. We thereby show that even language modeling can benefit from grounding, and the masked language modeling perplexity might translate to yet further applications in text generation \citep{wang-cho-2019-bert}. We will now investigate how grounding translates to downstream classification tasks accuracy.

\subsection{Downstream evaluation}

\begin{table*}
\centering
\begin{footnotesize}
\begin{tabular}{llllllllll}
\toprule
                            Strategy & Pretraining Text & K &           JOCI &           PDTB &            RTE &           SICK &            VUA &           VUAN &            AVG \\
\midrule
                              Albert-base &                   - &        - &            6.7 &           49.6 &  \textbf{68.0} &  \textbf{88.2} &           82.7 &           85.0 &           63.4 \\
                       
\midrule
                 No-Grounding &              Wiki-BK &        0 &            8.7 &           46.0 &           56.8 &           87.4 &           82.2 &           83.8 &           60.8 \\
                     Transferred i2t &        CC-SBU+Wiki-BK &      0/1 &           -0.4 &           31.1 &           60.2 &           85.7 &           82.1 &           84.5 &           57.2 \\
                     Transferred t2i &        CC-SBU+Wiki-BK &      0/1 &            8.3 &           19.3 &           65.6 &           87.4 &           82.9 &           84.6 &           58.0 \\
                 Transferred t2i+i2t &        CC-SBU+Wiki-BK &      0/1 &  \textbf{11.1} &  \textbf{52.0} &           61.1 &           88.0 &           82.7 &  \textbf{85.2} &           63.4 \\
 Associative \citep{Zhang2020Neural} &              Wiki-BK &       16 &            2.4 &           48.4 &           54.3 &           87.6 &           79.6 &           81.9 &           59.0 \\
           Associative (scene-based) &              Wiki-BK &       16 &            1.2 &           49.3 &           54.5 &           86.1 &           83.0 &           84.0 &           59.7 \\
          Associative (object-based) &              Wiki-BK &       16 &            8.6 &           50.8 &           67.6 &           87.1 &  \textbf{83.1} &           84.0 &  \textbf{63.5} \\
\bottomrule
\end{tabular}

\caption{Downstream tasks transfer results. Reported score is Spearman's correlation percentage for the JOCI task and accuracy percentage, otherwise.
Albert-base 
is the pretrained model from \citep{lan2019albert} without architectural change or further pretraining before the downstream fine-tuning.
Our models are based on Albert-base but have an additional cross-modal transformer that underwent an additional pretraining stage. $K$ is the number of images per example. The No-Grounding setup is equivalent to the transferred setup on Wiki-BK, or the associative setup with $K=0$. 
\label{tab:downstream}
}
\end{footnotesize}
\end{table*}

We hypothesize that grounded models should be better equipped to perform common sense related downstream tasks. A first component of our evaluation on tasks RTE, PDTB and JOCI will test that claim. But we also expect the concreteness to have an influence on the behavior of grounded models. Thus, we also perform a more targeted evaluation on concrete-only examples (SICK) and metaphoricity classification (VUA tasks) to better interpret our results.

\paragraph{Recognizing Textual Entailment (RTE)} \citep{rte} is a Natural Language Inference (NLI) task. Its dataset gathers sentence pairs with a premise and a hypothesis. The labels describe the logical relationship between the two, that is, entailment and non-entailment. 
\vspace{-0.2cm}

\paragraph{JHU Ordinal Common-sense Inference (JOCI)} \citep{zhang-etal-2017-ordinal} The dataset for this task also consists of premise/hypothesis sentence pairs, but the labels are numerical scores from 1 to 5 that reflect the plausibility of the hypothesis given the premise according to human annotators, relying on their own common sense.
\vspace{-0.2cm}

\paragraph{Sentences Involving Compositional Knowledge (SICK)} \citep{kiela-etal-2018-learning} is also a NLI dataset with entailment, neutral and contradiction classes but it differs from the previous two in that its premises are only composed of image captions and video descriptions. This allows a more specific evaluation of concrete language, as opposed to abstract domain language in the RTE tasks.
\vspace{-0.2cm}

\paragraph{Penn Discourse TreeBank (PDTB)} \citep{Prasad2014} contains a collection of fine-grained implicit (i.e., not signaled by a discourse marker) relations between sentences from the news domain in PDTB2.0, which signal the purpose of an utterance given a context utterance. We select level 2 implicit relations as categories. The task involves presupposition recognition and the ability to deal with non-literal meaning.
\vspace{-0.2cm}

\paragraph{VU Amsterdam Metaphor Corpus (VUA)} \citep{krennmayr2017vu} annotates the uses of verbs in sentences of the British National Corpus according to their level of metaphoricity. For instance, in the sentence {\it The alligator’s teeth are like white daggers}, the use of the word {\it daggers} is metaphorical while {\it teeth} is not. We use the dataset of a shared task on verb metaphoricity detection \citep{fig-lang-2020-figurative} as well as another version where we kept only the nouns which we call VUAN.

\vspace{0.0cm}

Table \ref{tab:sizes} in appendix \ref{sec:downstreamtasksizes} shows the dataset sizes, and Table \ref{tab:downstream} reports the results of the Wiki-BC trained models on the above tasks. We perform $8$ runs for each task and report the median score.


As baselines we propose the Albert-base model alongside an ablated model without visual grounding and a state-of-the-art association model which uses visual input as described by \citet{Zhang2020Neural}. The Albert-base model \citep{lan2019albert} is a pretrained transformer that takes the text tokens as inputs and provides a representation of the text examples at the output of the $[CLS]$ token which is used to perform the logistic regression. In the No-Grounding model, the cross-modal encoder is trained on text-only corpus always using the placeholder as visual input. \citep{Zhang2020Neural} is an image retrieval system that ranks images according to the number of text keywords their captions contain. In this work, the retrieved image features are combined with the source text features to perform neural machine translation. Here, we use this retrieval model with our associative strategy as a baseline.



The No-Grounding model performance does not match the Albert-base model, which indicates that our linguistic pretraining is not as well tuned. However, grounding still yields improvements on the JOCI, PDTB and VUA tasks. The image-to-text (i2t) pretraining does not transfer well in the absence of images. But combining both text-to-image and image-to-text training leads to higher results. This suggests that the resulting model is able to use the obtained multimodal features 
to better perform the downstream tasks. The performance of the associative strategies depends on the image retrieval system but the comparison suggests that object-based and scene-based retrieval perform well enough to yield meaningful results. 

Overall, the object-based associative strategy has the best performance and also performs well on language modeling, especially when generating descriptions.
But the transferred strategy seems to be a better choice for the downstream NLP tasks, since it does not require images during fine-tuning and at test time.


\section{Related work}

Visual grounding has been repeatedly applied in NLP. The specificity of our work lies in the systematic categorization and comparison of grounding strategies, alongside the proposition of image-to-text and associative grounding. 

It was shown to improve machine text translation \citep{specia2016shared,elliott-etal-2017-findings}. However, these improvements only affected multimodal translation tasks. \citet{Zhang2020Neural} showed that text-only translation tasks could benefit from external visual knowledge through image search, which we also demonstrate in our associative grounding strategy when applied to different NLP tasks. 

Visual grounding has also been applied to transferable word representation learning. \citet{bruni2014multimodal} generalize the distributional hypothesis by extracting discrete visual words from images and using them as context of text while relying on dimensionality reduction.
This latter idea is reused with an autoencoder \citep{silberer-lapata-2014-learning} and a SkipGram model \citep{lazaridou-etal-2015-combining}. \citet{kiela-bottou-2014-learning} combine word embeddings with ImageNet visual features associated to words. Conversely, \citet{ collell2017imagined} learn to predict ImageNet visual features from words and use the predicted {\it imagined} visual representation as auxiliary features.
However, the above techniques are not applicable beyond the word-level. Image captioning datasets do provide sentences associated with relevant images, but the multimodal models trained on these are not commonly used in downstream NLP tasks.
\citet{kiela-etal-2018-learning} train sentence embeddings for text processing tasks by learning to predict images from captions and report marginal improvements on SentEval \citep{conneau-kiela-2018-senteval} downstream tasks. This approach is similar to our text-to-image transferred grounding.
Concurrent work incorporate grounding into Transformers-based pretrained language models \citep{tan-bansal-2020-vokenization}. They match each word in a text-only corpora with an image, and perform a masked image modeling with associated tokens. In our framework, this could be Transferred Associative object-based Grounding.

Other work targets grounded language understanding, but mostly in the context of robotics \citep{matuszek2018grounded} where the text mostly regards task-oriented interactions in a closed-world.

Numerous text-level encoders were recently proposed to leverage images but were only applied to tasks involving both text and images. They all take BERT \citep{devlin-etal-2019-bert} as a starting point and represent images with region features. They adapt masked language modeling to images with a form of masked image modeling, and each have more specific contributions. For instance, LXMERT \citep{tan-bansal-2019-lxmert} demonstrates the value of visual question answering as a transferable pretraining task. UNITER \citep{chen2019uniter} shows that masked image modeling and masked language modeling are best done separately. VL-BERT \citep{su2019vl} uses an additional text-only training. VilBERT \citep{lu2019vilbert} proposes a KL-divergence loss for masked image modeling. 

Ideas comparable to associative visual grounding have also previously been used with external graph and text data. \citet{weijie2019kbert} leverage triplets found in knowledge bases to better find entities in text.
\citet{guu2020realm} use language modeling pretraining in order to learn to retrieve and use relevant passages in question answering tasks.

\section{Conclusion}
In this paper we have proposed visual grounding strategies to make joint text-image models applicable to text-only processing. We have demonstrated that the associative strategy leads to consistent improvements when performing NLP tasks such as masked language modelling, plausibility estimation, metaphoricity detection and discourse relation prediction. Results could be further improved by refining the image representation and retrieval model. Since relying on image captions limited the number of images we could use, it would be interesting to investigate the use of larger-scale image datasets. Further work is needed to refine the effect of multimodality on NLP tasks. It is also interesting to study how NLP performance scales with the sizes of datasets used in pretraining the multimodal representations  as is also suggested in \citep{kaplan2020scaling}.

\section{Acknowledgements}
This work is part of the CALCULUS5 project, which is
funded by the ERC Advanced Grant H2020-ERC-2017­
ADG 788506\footnote{\url{https://calculus-project.eu/}}.

\bibliographystyle{acl_natbib}
\bibliography{main}

\clearpage

\appendix 
\section{Appendix}\label{sec:appendix}

\subsection{Image association systems}

\begin{figure}[H]
\vspace{-0.3cm}
  \centering
\includegraphics[width =0.49\textwidth]{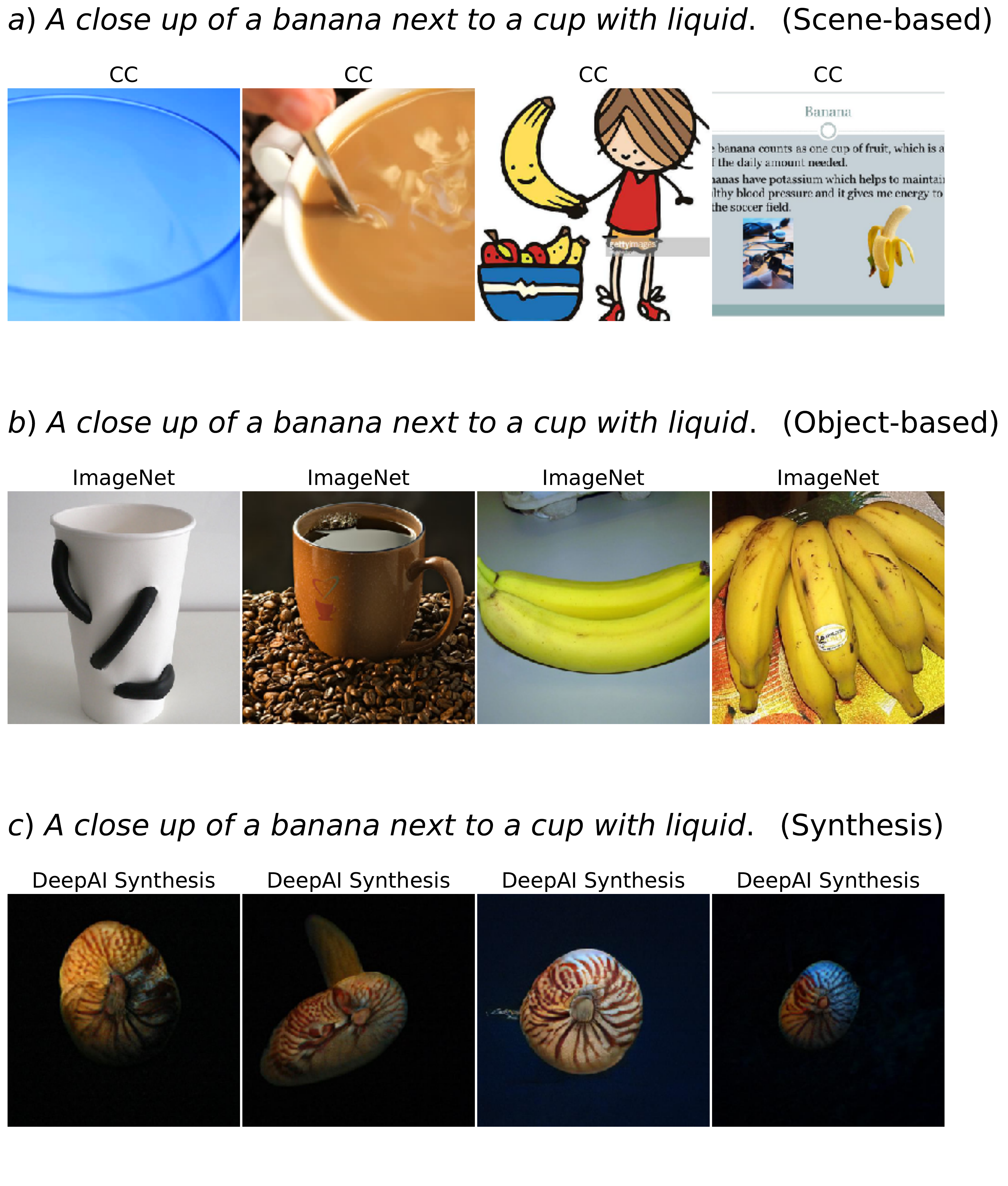}

\vspace{-0.5cm}
 
  \caption{Association of images with a random text sample from the COCO dataset, with scene-based and object-based matching with $K=4$ and $\kappa=2$. Scene-based matching $(a)$ indexes images by their caption embedding to match it with the input text. CC refers to the Conceptual Captions dataset.
   Object-based matching  $(b)$ identifies key concepts and matches them with ImageNet images. The Synthesis $(c)$ example illustrates the failure of the DeepAI image synthesis system when the input text is out of domain.
  }
\label{fig:matching} 
\end{figure}

\subsection{Downstream tasks sizes}\label{sec:downstreamtasksizes}
\begin{table}[H]
\begin{normalsize}
\centering
\begin{tabular}{lll}
\toprule
 Task & \#Examples (Train/Val/Test) & \#Labels \\
\midrule
 JOCI &                2.4k/299/298 &       - \\
  RTE &                2.2k/249/277 &       2 \\
 SICK &               4.5k/500/4.9k &       3 \\
 PDTB &             12.9k/1.2k/1.1k &      16 \\ 
VUAN &            5.0k/1.3k/2.2k &       2 \\
  VUA &             16k/1.7k/5.9k &       2 \\
\bottomrule
\end{tabular}
\caption{Dataset sizes for each downstream task\label{tab:sizes}
}
\vspace{-0.3cm}
\end{normalsize}
\end{table}

\end{document}